\documentclass[sn-mathphys, Numbered]{sn-jnl}

\usepackage{graphicx}
\usepackage{multirow}
\usepackage{amsmath,amssymb,amsfonts,bm}
\usepackage{amsthm}
\usepackage{mathrsfs}
\usepackage[title]{appendix}
\usepackage{xcolor}
\usepackage{textcomp}
\usepackage{manyfoot}
\usepackage{booktabs}
\usepackage{algorithm}
\usepackage{algorithmicx}
\usepackage{algpseudocode}
\usepackage{subcaption}
\usepackage{listings}
\usepackage{latexsym}
\usepackage{url}
\usepackage{arydshln}
\usepackage{tabulary}
\usepackage{epstopdf}

\newcommand{\revision}[1]{\textcolor{black}{#1}} 

\begin{document}

\title{
ConVibNet: Needle Detection during Continuous Insertion via Frequency-Inspired Features
}

%

\author{\fnm{Jiamei} 
\sur{Guo$^{\dagger,1}$}}
\email{jiamei.guo@tum.de}

\author{\fnm{Zhehao} 
\sur{Duan$^{\dagger,1}$}}
\email{zhehao.duan@tum.de}

\author{\fnm{Maria} 
\sur{Neiiendam$^{1,3}$}}
\email{maria.neiiendam@gmail.com}

\author*{\fnm{Dianye} 
\sur{Huang$^{*,1,2}$}}
\email{dianye.huang@tum.de}

\author{\fnm{Nassir} 
\sur{Navab$^{1,2}$}}
\email{nassir.navab@tum.de}

\author{\fnm{Zhongliang} 
\sur{Jiang$^{1,2,4}$}}
\email{zljiang@hku.hk}





\affil[1]{Computer Aided Medical Procedures and Augmented Reality (CAMP), Technical University of Munich, Munich, Germany}
\affil[2]{Munich Center for Machine Learning (MCML), Munich, Germany}
\affil[3]{Technical University of Denmark, Kongens Lyngby, Denmark}
\affil[4]{The University of Hong Kong, Hong Kong, China}
\affil[$\dagger$]{Equal Contribution}

\abstract{



\textbf{Purpose:}  
Ultrasound-guided needle interventions are widely used in clinical practice, but their success critically depends on accurate needle placement, which is frequently hindered by the poor and intermittent visibility of needles in ultrasound images. \revision{Existing approaches remain limited by artifacts, occlusions, and low contrast, and often fail to support real-time continuous insertion.} To overcome these challenges, this study introduces a robust real-time framework for continuous needle detection.

\textbf{Methods:}
We present ConVibNet, an extension of VibNet for detecting needles with significantly reduced visibility, \revision{addressing real-time, continuous needle tracking during insertion.} ConVibNet leverages temporal dependencies across successive ultrasound frames to enable continuous estimation of both needle tip position and shaft angle in dynamic scenarios. To strengthen temporal awareness of needle-tip motion, we introduce a novel intersection-and-difference loss that explicitly leverages motion correlations across consecutive frames. \revision{In addition, we curated a dedicated dataset for model development and evaluation.}

\textbf{Results:}
The performance of the proposed ConVibNet model was evaluated on our dataset, demonstrating superior accuracy compared to the baseline VibNet and UNet-LSTM models. Specifically, ConVibNet achieved a tip error of 2.80±2.42 mm and an angle error of 1.69±2.00°. These results represent a 0.75 mm improvement in tip localization accuracy over the best-performing baseline, while preserving real-time inference capability.

\textbf{Conclusion:}
ConVibNet advances real-time needle detection in ultrasound-guided interventions by integrating temporal correlation modeling with a novel intersection-and-difference loss, thereby improving accuracy and robustness and demonstrating high potential for integration into autonomous insertion systems.

}


\maketitle

\section{Introduction}
\par
Ultrasound-guided needle intervention is among the most widely practiced techniques for medical procedures such as biopsies, regional anesthesia, and ablations, due to the real-time and radiation-free nature of ultrasound (US) imaging~\cite{kuang2016modelling}. However, its efficacy heavily relies on the accurate localization of the inserted needle under the skin.
Automated needle detection in US images remains challenging due to the intrinsic limitations of speckle noise and low contrast, causing reduced needle visibility~\cite{huang2025vibnet}. 
This challenge is further intensified by the narrow imaging plane of US images and the needle’s susceptibility to occlusion from needle-like artifacts, resulting in intermittent visibility and posing significant obstacles to automated detection~\cite{zhang2025mambaxctrack, goel2023motion}.

Classical methods for needle detection in ultrasound imaging primarily relied on handcrafted features, such as statistical filters~\cite{mathiassen2016robust} and Gabor filters~\cite{kaya2015real}, along with geometric modeling to exploit the distinct appearance of the needle shaft and tip. With the recent advances in convolutional neural networks (CNNs), data-driven approaches have demonstrated superior robustness and generalization~\cite{jiang2023robotic, bi2024machine, jiang2024intelligent}. Representative examples include the use of fully convolutional networks combined with region-based CNNs for needle trajectory estimation~\cite{mwikirize2018convolution}, and methods incorporating adaptive compression and Radon-transform-based segmentation for automated guidance in core-needle biopsy procedures~\cite{wijata2021automatic}. Although these works have shown satisfactory results, they still face challenges, particularly when relying on single-frame information, which makes them susceptible to occlusions, acoustic artifacts, and low-contrast imaging conditions.

To address these challenges, temporal information has been incorporated into detection frameworks to exploit motion cues and improve tracking stability~\cite{huang2023motion}. Representative approaches include architectures that integrate CNNs with long short-term memory (LSTM) modules~\cite{belikova2021deep} and encoder–decoder networks augmented with Kalman filter-inspired components to capture motion continuity and reduce tip localization errors~\cite{goel2023motion}. More recently, Zhang~\emph{et al.}~\cite{zhang2025mambaxctrack} introduced MambaXCTrack, which employs structured state space modeling enhanced by cross-correlation and implicit motion prompting for ultrasound-guided needle tracking, effectively mitigating image noise and intermittent tip invisibility. Nevertheless, approaches that rely solely on temporal variations in the needle’s visual appearance may suffer from limited generalizability when the background texture during inference differs from that encountered during training.

Hardware-augmented approaches include modifying the needle surface with features such as grooves, dimples, or specialized coatings to enhance visibility in US imaging, using echogenic needles~\cite{hovgesen2022echogenic}, or employing needles with a Doppler twinkling signature~\cite{dupere2023new}. However, these methods increase the cost of the needle and may introduce additional imaging artifacts. To mitigate dependence on the spatial texture of the input, a viable strategy is to examine the variations in pixel intensities across successive US images. 
Beigi~\emph{et al.}~\cite{beigi2016spectral} explored using manual tremor motion to aid needle detection in B-mode US, achieving better results with convex probes where the shaft is initially visible near the puncture site. The detection performance, however, degraded in in-vivo tests due to motion noise from surrounding tissues. To address tremor-induced artifacts in hand-held probes, they further proposed a spatiotemporal tracking framework~\cite{beigi2017detection}, but its reliance on handcrafted features and multi-scale spatial decomposition limited processing to one frame per second. More recently, Huang~\emph{et al.}~\cite{huang2025vibnet} developed VibNet, which induces mechanical vibration in the needle to generate periodic motion signatures that can be captured via frequency-domain feature extraction. Although VibNet improves detection robustness even when the needle is nearly invisible in B-mode imaging, it is restricted to static needle detection and cannot handle continuous insertion.

\par
To further tackle this challenge, this study investigates the detection of the moving needle from ultrasound images during live insertion procedures using frequency features.  To better capture temporal dynamics, we propose a novel intersection-and-difference loss that explicitly leverages motion correlations across consecutive frames, thereby enhancing temporal awareness and consistency. To mitigate the difficulty of annotating nearly invisible needles in US images, we also developed a data acquisition platform equipped with an NDI tracking system and a needle guide for automatic annotation, coupled with data quality check to ensure reliable ground-truth labels. Experimental results from comparisons with baselines and ablation studies demonstrate that ConVibNet achieves superior and real-time performance, validating the effectiveness of its design and the advantages of temporal motion modeling in the frequency domain. Overall, this work positions ConVibNet as a promising framework for ultrasound-guided needle detection, with high potential for integration into motorized insertion systems and future autonomous platforms.

\section{Methods}
\begin{figure}[t!]
  \centering
  \includegraphics[width=0.90\textwidth]{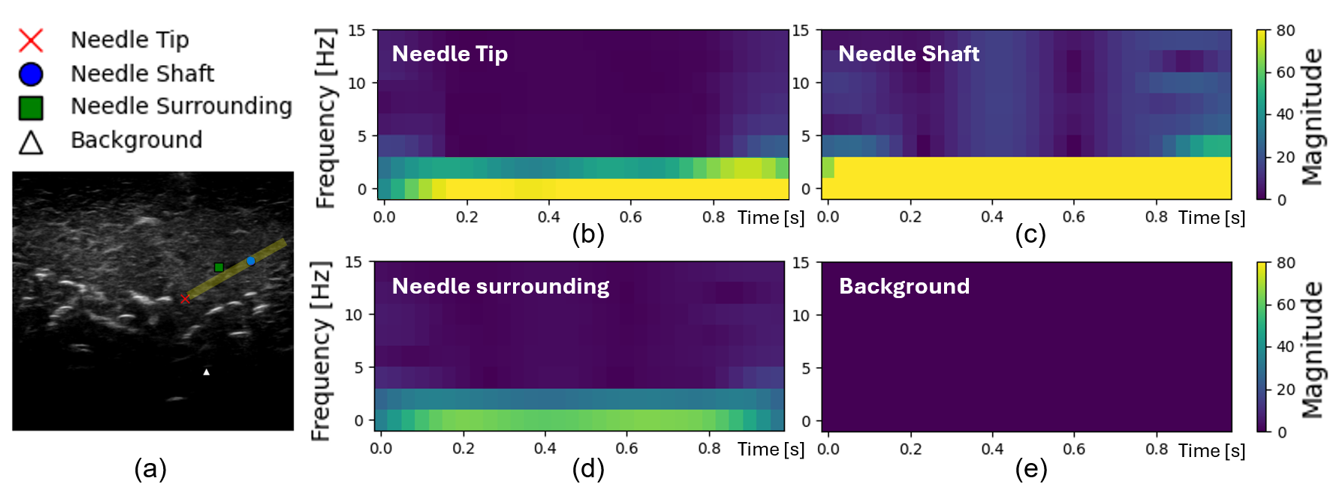}
  \caption{Frequency analysis of the US image sequence. (a) Selected pixel locations for analysis in the $t$-th frame, with the yellow transparent line indicating the needle at an insertion angle of 30°. (b)$\sim$(e) Spectrograms corresponding to the needle tip, shaft, surrounding tissue, and background.}
  \label{fig:motivation}
\end{figure}

This section presents ConVibNet, a framework designed to capture temporal dependencies across successive ultrasound frames for real-time needle detection. Specifically, ConVibNet extends VibNet to dynamic insertion scenarios, enabling continuous estimation of both the needle tip position and shaft angle. Such a task requires handling not only the intentionally applied small-amplitude vibrations but also additional insertion-induced needle motions, which may introduce confounding effects. It thereby raises a critical question: Does frequency-domain information remain sufficiently salient under insertion conditions? To investigate, we analyzed US sequences recorded at $30~\text{fps}$, where the needle was vibrated by the mechanism shown in Fig.~\ref{fig:expsetup}(a). As illustrated in Fig.~\ref{fig:motivation}, four representative points were manually selected within a 30-frame sequence: the needle tip, needle shaft, surrounding tissue, and background. Their temporal intensity variations were transformed using the Short-Time Fourier Transform (STFT) to generate time-frequency spectrograms. The results show that, compared with tissue [Fig.~\ref{fig:motivation}(d)] and background [Fig.~\ref{fig:motivation}(e)], the needle tip [Fig.~\ref{fig:motivation}(b)] and shaft [Fig.~\ref{fig:motivation}(c)] exhibit stronger and more distinct frequency components. Spectrogram magnitudes decrease progressively with distance from the shaft, becoming weakest in the surrounding tissue. These observations confirm that frequency-domain features remain prominent along the needle trajectory, which intuitively provides the basis for continuous needle detection.

\subsection{Intersection and Difference Loss for Motion Correlations}

\begin{figure}[t]
  \centering
  \includegraphics[width=0.95\textwidth]{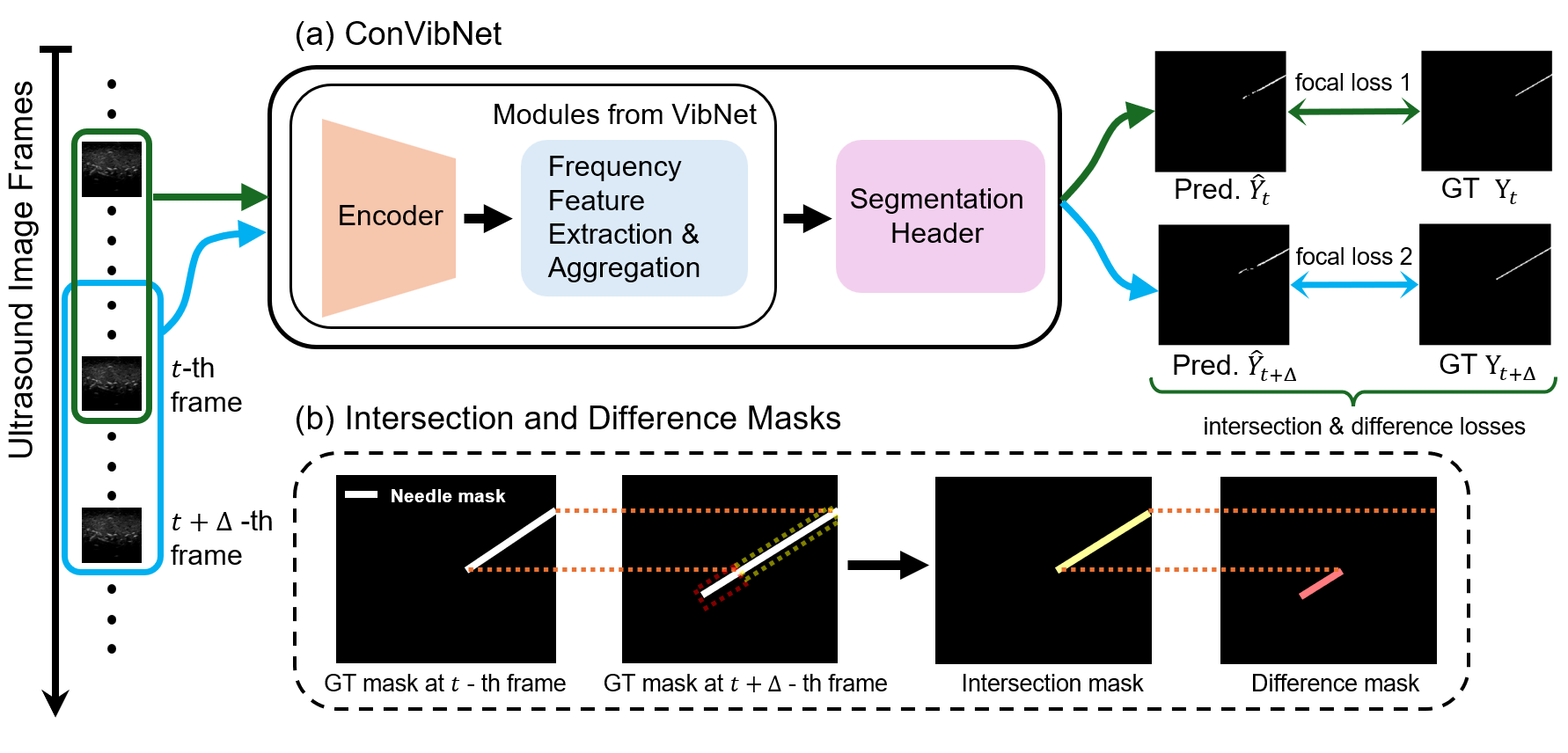}
  \caption{Overview of the proposed method. (a) Architecture of ConVibNet. (b) Illustration of how the intersection and difference masks are computed.}
  \label{fig:overview}
\end{figure}

Fig.~\ref{fig:overview} provides an overview of the proposed method. Extending VibNet to the continuous settings introduces two major challenges. First, real-time needle detection is hindered by the Hough Transform module, which, as reported in~\cite{huang2025vibnet}, incurs high computational cost and is therefore unsuitable for real-time tip tracking. To address this, we retain only the motion encoder together with the frequency feature extraction and aggregation modules from VibNet~\cite{huang2025vibnet}, while replacing the deep Hough Transform (DHT) with a segmentation head [see Fig.~\ref{fig:overview}(a)]. Second, to alleviate class imbalance, we employ focal loss~\cite{Lin2017}, as defined in Eq.~(\ref{eq:focal_loss}). This loss suits well for thin needle detection~\cite{jiang2024needle}. It adaptively down-weights well-classified examples, reducing the influence of abundant negative samples and emphasizing positive ones, thereby guiding the model to focus more effectively on the needle mask.
\begin{equation}
\mathcal{L}_{f} = -\frac{1}{N} \sum_{i} \Big[ 
    y_i \left(1 - \hat{y}_i\right)^\eta \log\left(\hat{y}_i\right)
    + \left(1 - y_i\right) \left(1 - \hat{y_i}\right)^\gamma \hat{y}_i^\eta \log\left(1 - \hat{y}_i\right)
\Big] 
\label{eq:focal_loss}
\end{equation}
where $\eta=2$ and $\gamma=4$; $y_i$ and $\hat{y}_i$ denote the ground truth and the predicted output.

Secondly, VibNet processes a sequence of $L$ consecutive frames but yields only a single segmentation output $Y$. In contrast, our task involves continuous needle insertion, where segmentation is required specifically for the final frame of each sequence. Incorporating frequency-domain features introduces temporal aggregation, which complicates isolating information relevant solely to this last frame. To overcome this, we augment the loss function to guide ConVibNet in capturing temporal motion correlations by comparing predictions from two input sequences whose final frames are separated by $\Delta$ time steps. In our implementation, each sequence consists of $L=30$ frames with $\Delta=5$.

As illustrated in Fig.~\ref{fig:overview}, the two sequences are processed independently by ConVibNet, and the focal losses [see Eq.~(\ref{eq:focal_loss})] are computed for both outputs. We then introduce two auxiliary loss terms: intersection loss ($\mathcal{L}_{\text{inter}}$) and difference loss ($\mathcal{L}_{\text{diff}}$). The intersection loss is designed to enhance the model’s accuracy in fine-grained regions by emphasizing consistent predictions, whereas the difference loss encourages the model to capture temporal dynamics between the sequences, thereby improving its ability to learn time-dependent features. Fig.~\ref{fig:overview} (b) illustrates how the intersection and difference masks are computed between two ground truth or predicted segmentation outputs. 

The total loss function of ConVibNet is formulated as a weighted sum of four components: the focal losses for each input sequence, the intersection loss, and the difference loss.
\begin{equation}
\label{eq:loss_fn}
\mathcal{L} = \mathcal{L}_{f}^{(t)} + \mathcal{L}_{f}^{(t+\Delta)} + \alpha\mathcal{L}_{\text{inter}} + \beta\mathcal{L}_{\text{diff}}
\end{equation}
where \[
    \mathcal{L}_{\text{inter}} = \text{BCE }\left(\hat{Y}_{t} \cdot \hat{Y}_{t+\Delta},\ {Y}_t \cdot {Y}_{t+\Delta}\right),~~
    \mathcal{L}_{\text{diff}} = \text{BCE }\left(|\hat{Y}_t - \hat{Y}_{t+\Delta}|,\ |Y_t - Y_{t+\Delta}|\right), \]
$\alpha$ and $\beta$ are hyperparameters that control the contribution of the intersection and difference losses, respectively; ``~$\cdot$~'' denotes element-wise multiplication. 

Notably, the difference loss is introduced during training only after the model reaches a relatively stable state. Empirically, applying it too early can impede convergence, likely because enforcing dynamic consistency before the model has learned basic representations adds excessive complexity. In this work, the difference loss was activated from the second epoch onward. The model was optimized using the Adam algorithm with a learning rate of \(1 \times 10^{-4}\) \revision{and a batch size of 4}. To prevent overfitting, we employed early stopping with training typically converging by the 3\textsuperscript{rd} epoch, \revision{corresponding to a small number of full passes over the training set given the dataset size described below}. Since the predicted needle segmentation mask can be discontinuous, we applied the RANSAC algorithm 
during post-processing to fit a line and improve robustness. Finally, the positive pixel closest to the bottom of the image is projected onto the fitted line to determine the needle tip position.

\subsection{Data Preparation}

\subsubsection{Experimental Setup}
\label{sec:expsetup}
\par
\begin{figure}[t!]
  \centering
  \includegraphics[width=0.98\textwidth]{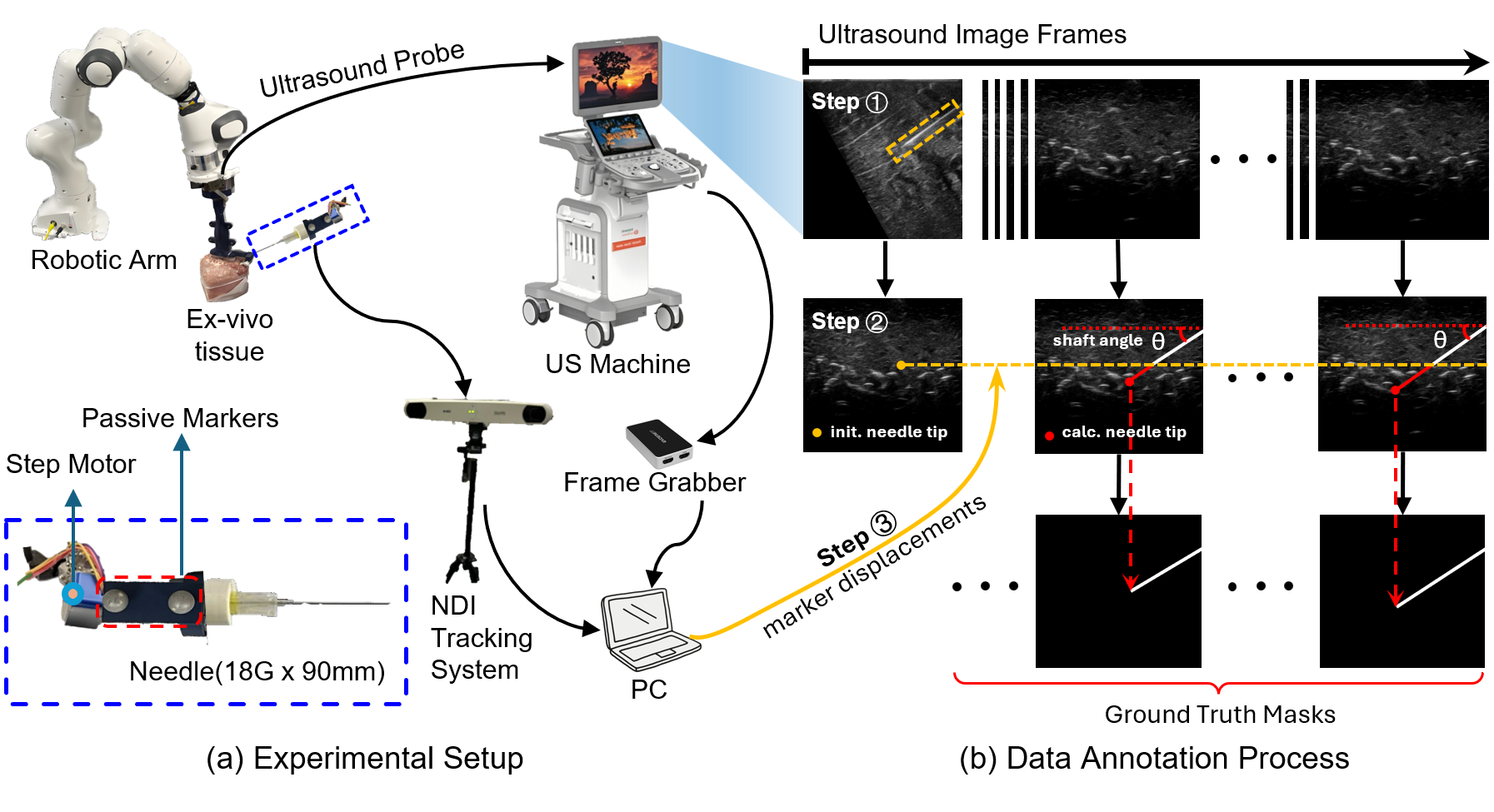}
  \caption{Overview of the data acquisition workflow. (a) Hardware setup and connections (see \textit{Sec.~\ref{sec:expsetup}}). (b) US image sequence annotation process (see \textit{Sec.~\ref{sec:gtanno}}).}
  \label{fig:expsetup}
\end{figure}

A major challenge in US imaging is the poor visibility of the needle, particularly during tissue penetration, which makes direct annotation of the needle tip as ground truth difficult. To overcome this, we built a data acquisition platform [Fig.\ref{fig:expsetup} (a)] using ex vivo porcine tissue to simulate realistic soft tissue conditions. To increase dataset diversity, insertion locations were randomly varied. An $18$-gauge, $90~mm$ needle was inserted along a guide at one of two predefined angles ($\theta=15^\circ$ or $30^\circ$) to ensure consistent trajectories, \revision{covering both a shallow insertion case and a relatively steeper scenario under the constraints of the experimental setup}. During insertion, the needle shaft was vibrated at approximately $2.5~Hz$ by a stepper motor (28BYJ-48-5V, Pollin Electronic GmbH, Germany) driven by a motor controller, introducing temporal frequency components for downstream analysis. A 3D-printed connector transmitted vibrations from the motor shaft to the needle hub, while two passive markers mounted on the connector enabled real-time optical tracking of needle displacement at $100~Hz$ using the NDI system (Polaris Vicra, Northern Digital Inc., Canada). US images were acquired at $30~fps$ with a Siemens ACUSON Juniper system (SIEMENS AG, Germany) using a linear probe (12L3, footprint $51.3~mm$) \revision{with an imaging depth of 4.5 cm}. The image stream was captured and transferred to a workstation in real time via a Magewell frame grabber (Magewell, China) for further processing.

\subsubsection{Data Acquisition and Annotation Process}
\label{sec:gtanno}
\par
As shown in Fig.~\ref{fig:expsetup}(b), needle mask annotation for each insertion video is performed in 3 steps. \textbf{Step 1}: With the US probe fixed by the robotic arm, the needle is inserted along a predefined needle guide, and the US beam angle is adjusted for optimal visibility. \textbf{Step 2}: Once the needle tip is visible, it is manually annotated in the current frame, after which the beam angle is reset. \textbf{Step 3}: During insertion, the NDI system tracks passive markers on the needle, and their 3D trajectories are synchronized with each US frame through spatial conversion~\cite{huang2024robot}. The relative displacement of the tip between frames is computed from marker positions, allowing the ground-truth tip location in subsequent frames to be derived from the initial manual annotation. This process yields accurate needle masks, even when the needle is indistinct in the US images.

\par
During each insertion trial, the needle tip was repeatedly advanced and partially retracted several times to mimic typical clinical probing and adjustment maneuvers. This process continued until the needle was fully withdrawn. The procedure was repeated $60$ times for each predefined insertion angle, with each trial performed at a different tissue surface position, yielding a total of $120$ US videos. During dataset curation, frames from \textbf{Step 1} and \textbf{Step 2}, when the US beam had not yet been reset to its default orientation, were discarded along with their corresponding masks. Despite the use of a needle guide, the actual needle path may deviate from the predefined trajectory due to mechanical interactions with soft tissue, where heterogeneous and viscoelastic properties impose uneven forces on the shaft, causing bending or deflection~\cite{Abolhassani2007}. To account for this, a manual quality check was conducted: for each video, the generated masks were visually inspected and compared against the approximate needle location. Although the needle is nearly invisible in US images, its coarse position can be inferred from surrounding tissue motion. Sequences showing substantial angular deviation between the annotation mask and this rough estimation were removed. As shown in Tab.~\ref{tab:dataset}, the collected dataset comprises 106 videos, split into training, validation, and test sets in proportions of 80\%, 10\%, and 10\%, respectively. During the training, the sequences were augmented \revision{on-the-fly} through horizontal flipping and adjustments to image contrast and brightness \revision{to improve robustness and mitigate overfitting given the limited dataset size}. Each sequence consists of $30$ consecutive US frames.

\begin{table}[b!]
\centering
\setlength{\tabcolsep}{8pt}
\caption{Overview of the Curated Dataset}
\label{tab:dataset}
\begin{tabular}{ccccc} 
\toprule
\textbf{Insertion Angle} & \textbf{Total} & \textbf{Training Set} & \textbf{Validation Set} & \textbf{Test Set} \\
\midrule
15° & 50/10380 & 40/8629 & 5/825 & 5/926 \\
30° & 56/22965 & 44/18157 & 6/2298 & 6/2510 \\
\bottomrule
\multicolumn{5}{l}{Values are presented in the format: \textit{num of videos/num of sequences}.} 
\end{tabular}
\end{table}

\section{Evaluation and Results}
\par
We systematically evaluate the proposed model against relevant baselines for dynamic needle detection. The experiments compare conventional intensity-based methods with our frequency-domain variants in terms of needle tip localization and insertion angle estimation. All models were trained on the same dataset, and performance was assessed using three metrics: \textit{(i) Tip error} (mm): mean Euclidean distance between predicted and ground-truth tip positions; \textit{(ii) Angle error} (°): mean absolute deviation between predicted and ground-truth shaft angles; \textit{(iii) Success rate} (\%): proportion of test samples with tip error $<10$ mm and angle error $<15^\circ$, indicating reliable needle detection.

\subsection{Comparison Study} 
\par
To validate the proposed method, we compare ConVibNet against two baselines: (i) VibNet w/o DHT, where the original DHT module is replaced with a segmentation head identical to ConVibNet for fair comparison, since the DHT incurs excessive computational overhead for real-time use; and (ii) UNet-LSTM~\cite{belikova2021deep}, which integrates UNet with an LSTM to capture temporal dependencies. VibNet w/o DHT serves as a frequency-domain baseline, while UNet-LSTM represents a conventional intensity-based spatio-temporal model. \revision{While specialized hardware solutions such as echogenic needles~\cite{hovgesen2022echogenic} or Doppler-based tracking~\cite{dupere2023new} exist, they often incur higher costs or require specific imaging modes.} As shown in Tab.~\ref{tab:cmpres}, ConVibNet achieves the lowest tip error of 2.80±2.42 mm, improving by 0.75 mm over VibNet w/o DHT and 0.80 mm over UNet-LSTM. While the angle error increases slightly to 1.69±2.00° compared to ~1.59° for both baselines, ConVibNet maintains comparable angular accuracy. Notably, the success rate reaches 79.6\%, compared to 63.7\% for VibNet w/o DHT, demonstrating superior overall reliability in dynamic needle detection. \revision{For inference, ConVibNet runs with an inference time of 33~ms per sample on an RTX 1080Ti GPU, matching the 30~Hz frame rate commonly used in clinical US imaging.}

\begin{table}[h!]
\centering
\setlength{\tabcolsep}{14pt}
\caption{Needle detection performance compared with baseline models}
\label{tab:cmpres}
\begin{tabular}{@{}lccc@{}} 
\toprule
\textbf{Methods} & \textbf{Tip Err. (mm)} & \textbf{Angle Err. (°)} & \textbf{Suc. Rate (\%)} \\
\midrule
VibNet~\cite{huang2025vibnet} w/o DHT & $3.55 \pm 2.65$ & $1.59 \pm 2.13$ & $63.7$ \\
UNet-LSTM~\cite{belikova2021deep} & $3.60 \pm 2.75$ & $\mathbf{1.59 \pm 1.78}$ & $62.7$ \\
ConVibNet$^\dagger$ (ours) & $\mathbf{2.80 \pm 2.42}$ & $1.69 \pm 2.00$ & $\mathbf{79.6}$\\
\bottomrule
\multicolumn{4}{l}{The angle error and tip error are described in the format of mean$\pm$std.} \\
\multicolumn{4}{l}{$^\dagger$Parameter settings for the proposed ConVibNet: $\alpha = 0.5$ and $\beta = 0.02$.} \\
\multicolumn{4}{l}{Err.: error;~~Suc. success.}\\
\end{tabular}
\end{table}

\par
\begin{figure}[h]
  \centering
  \includegraphics[width=0.9\textwidth]{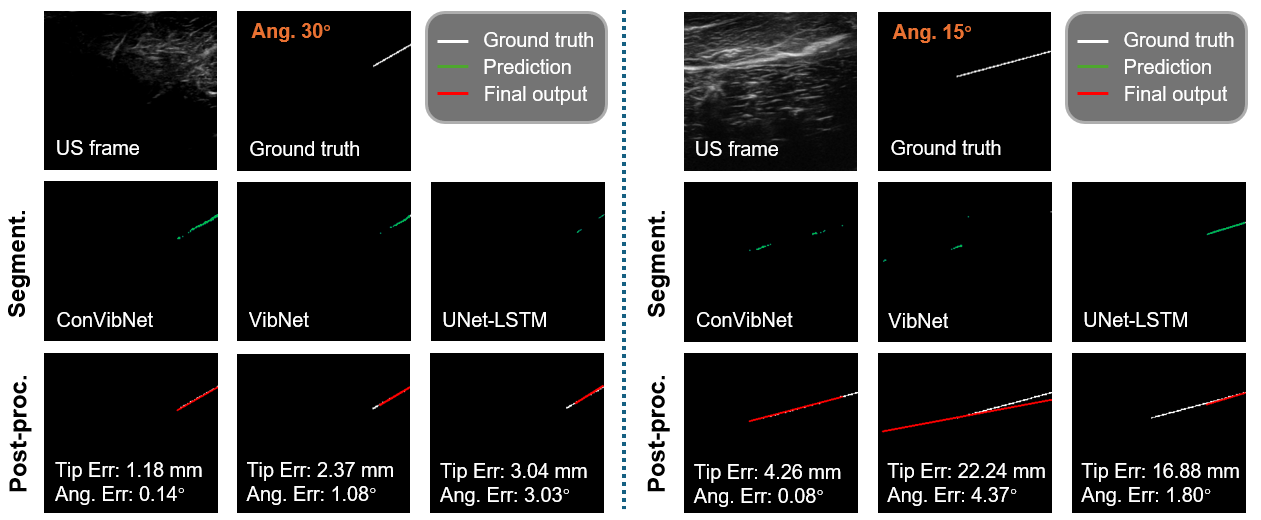}
  \caption{Example comparison of predicted segmentation and post-processed outputs}
  \label{fig:cmpres}
\end{figure}

The qualitative results in Fig.~\ref{fig:cmpres} illustrate the improvements achieved by ConVibNet across different insertion angles. At $30^\circ$, ConVibNet generates more continuous segmentation masks that accurately capture the needle tip, whereas baseline methods produce fragmented shaft detections, leading to large tip errors. Similar improvements are observed at $15^\circ$, a more challenging case \revision{where the needle is surrounded by nearby elongated tissue structures with needle-like appearances. These structures increase motion pattern ambiguity and consequently degrade tracking performance}. Here, VibNet yields excessive and noisy detections that overestimate needle length after post-processing, while UNet-LSTM detects only a short segment near the entry point, missing the deeper tip and underestimating length. In contrast, ConVibNet maintains robust segmentation and more accurate tip localization under both conditions, even in the challenging visibility scenarios.

\subsection{Ablation Study}
\par
To evaluate the distinct contributions of the proposed loss functions in~Eq. (\ref{eq:loss_fn}), we conducted systematic ablation studies, with results summarized in Table~\ref{tab:ablres}. \revision{Unless otherwise stated, all ablation experiments were performed using the same network architecture, training schedule, data splits, and augmentation strategy, with only the loss configuration varied.} The reference framework (``w/o $\mathcal{L}_{\text{inter}}$ and $\mathcal{L}_{\text{diff}}$"), trained without the intersection and difference loss terms, established a performance benchmark with a tip error of $3.55 \pm 2.65$ mm, an angle error of $1.59 \pm 2.13$° and a success rate of $63.7\%$.

We first examine the impact of the intersection loss controlled by $\alpha$ [see Eq. (\ref{eq:loss_fn})], which is designed to enhance accuracy in fine-grained regions by enforcing consistency across overlapping input sequences (see ``w/ $\mathcal{L}_{\text{inter}}$, w/o $\mathcal{L}_{\text{diff}}$"). Introducing the intersection loss demonstrated a substantial performance improvement and revealed its dominant role. The optimal result was achieved at $\alpha=0.50$, where the tip error significantly decreased to $2.93 \pm 2.54$ mm, and the success rate surged to $81.4$\%. However, a lower weight ($\alpha=0.25$) or a higher weight ($\alpha=0.75$) resulted in worse performance, underscoring the need for careful tuning. 

With $\alpha$ fixed at its optimal value of $0.50$, we then analyzed how the difference loss, governed by $\beta$ [see Eq. (\ref{eq:loss_fn})], contributes to overall performance by systematically varying $\beta$. The best overall performance was achieved with the combination of $\alpha=0.50$ and $\beta=0.02$, which yielded the lowest tip error of $2.80 \pm 2.42$ mm and an angle error of $1.69 \pm 2.00$°. This suggests that the difference loss acts as an effective regularizer that encourages the model to capture temporal dynamics. Crucially, when the difference loss was used in isolation (see ```w/o $\mathcal{L}_{\text{inter}}$, w/ $\mathcal{L}_{\text{diff}}$"), performance degraded significantly, with tip error increasing to $4.04 \pm 2.71$ mm. This shows that the difference loss works not on its own, but as a useful regularizer that makes use of spatially consistent features.

In summary, the ablation studies show that both loss components play important and complementary roles. The intersection loss mainly improves performance by maintaining spatial consistency, while the difference loss serves as a temporal regularizer that refines the results over time. The best performance is achieved with $\alpha$=0.5 and $\beta$=0.02, which provides a good balance between accuracy and robustness. These findings confirm the effectiveness of the proposed intersection-and-difference loss design.

\begin{table}[t!]
\caption{Ablation study on the intersection and difference loss}
\label{tab:ablres}
\centering
\tabcolsep=0.25cm
\begin{tabular}{lllccc}
\toprule
\multicolumn{3}{c}{\textbf{ConVibNet}} & \multirow{2}{*}{\textbf{Tip Err. (mm)}} & \multirow{2}{*}{\textbf{Angle Err. (°)}} & \multirow{2}{*}{\textbf{Suc. Rate (\%)}} \\
\cmidrule(l){1-3} 
\multicolumn{1}{c}{\textbf{Frameworks}} & \multicolumn{1}{c}{$\alpha$} & \multicolumn{1}{c}{$\beta$} & & & \\
\midrule
\multirow{1}{*}{w/o $\mathcal{L}_{\text{inter}}$ and $\mathcal{L}_{\text{diff}}$}
 & $0.00$ & $0.00$ & $3.55 \pm 2.65$ & $1.59 \pm 2.13$ & $63.7$ \\
\cmidrule(l){2-6} 
\multirow{3}{*}{w/ $\mathcal{L}_{\text{inter}}$, w/o $\mathcal{L}_{\text{diff}}$}
 & $0.25$ & $0.00$ & $3.66 \pm 2.85$ & $1.87 \pm 2.34$ & $66.9$ \\
 & $0.50$ & $0.00$ & $2.93 \pm 2.54$ & $1.78 \pm 2.24$ & $\mathbf{81.4}$ \\ 
 & $0.75$ & $0.00$ & $3.49 \pm 2.90$ & $1.86 \pm 2.24$ & $77.0$ \\
 \cmidrule(l){2-6} 
 \multirow{3}{*}{w/ $\mathcal{L}_{\text{inter}}$ and $\mathcal{L}_{\text{diff}}$}
 & $0.50$ & $0.01$ & $2.83 \pm 2.45$ & $1.76 \pm 2.12$ & $80.5$ \\
 & $\mathbf{0.50}$ & $\mathbf{0.02}$ & $\mathbf{2.80 \pm 2.42}$ & $\mathbf{1.69 \pm 2.00}$ & $79.6$ \\ 
 & $0.50$ & $0.04$ & $2.94 \pm 2.86$ & $1.73 \pm 2.21$ & $69.2$ \\
 \cmidrule(l){2-6} 
\multirow{1}{*}{w/o $\mathcal{L}_{\text{inter}}$, w/ $\mathcal{L}_{\text{diff}}$}
 & $0.00$ & $0.02$ & $4.04 \pm 2.71$ & $2.01 \pm 2.51$ & $58.3$ \\ 
\bottomrule
\multicolumn{6}{l}{The angle and tip error are described in the format of mean$\pm$std; 
}\\
\multicolumn{6}{l}{Err.: errors;~~Suc.: success;~~w/o: without;~~w/: with.}
\end{tabular}
\end{table}

\section{Discussion and Conclusion}
\par
The proposed ConVibNet achieves real-time performance in continuous needle detection \revision{at typical clinical ultrasound frame rates.} Along with this computational efficiency, ConVibNet delivers superior needle-tip localization accuracy ($2.80\pm2.42~mm$) and a success rate of 79.6\%, representing a 0.75 mm gain in tip accuracy and a 15.9\% improvement in success rate compared to the best-performing baseline. These results confirm the effectiveness of the proposed intersection-difference loss in enhancing temporal awareness of needle motion, leading to robust performance under challenging imaging conditions.

\par
Despite these promising outcomes, several limitations remain. First, data collection is constrained by the difficulty of acquiring high-quality annotations, even with the aid of an NDI tracking system, and by the limited range of insertion angles (15° and 30°) available in the current setup, \revision{which were selected to ensure repeatable trajectories rather than to exhaustively cover all clinically possible insertion angles}. As a result, the performance of shaft-angle prediction could not be fully evaluated. Additionally, needle bending and the effect of operator variability were not explicitly addressed in this work. \revision{Furthermore, generalization to other probe types, US systems, and needle specifications was not validated.} Future research will therefore investigate more diverse insertion settings, incorporate needle bending into the detection framework, \revision{assess the influence of human-induced variability, and validate generalization across different imaging configurations and needle specifications}. Building a robotic platform equipped with optical fiber sensors could further facilitate reliable ground-truth acquisition by simultaneously capturing both needle pose and deformation during insertion. A further methodological limitation lies in the model’s implicit handling of motion dynamics. The observed needle movement arises from two sources: operator manipulation and superimposed high-frequency vibration for signal enhancement. Since the current model does not disentangle these components, its ability to learn a fully robust representation may be restricted. Future architectures capable of explicitly separating these motion sources could yield more accurate and generalizable predictions. 

\par
In conclusion, we present ConVibNet, an extension of VibNet tailored for continuous needle detection in ultrasound-guided interventions. Through its novel intersection-difference loss design and temporal correlation modeling, ConVibNet achieves significant gains in both accuracy and robustness while maintaining real-time performance. These advances highlight its potential as a viable candidate for integration into motorized needle insertion systems and as a promising step toward autonomous ultrasound-guided procedures.

\section*{Declarations}
\subsection*{Conflict of Interest}
\revision{The authors declare that they have no conflict of interest.}

\subsection*{Funding}
\revision{This work was supported by the Multi-Scale Medical Robotics Center, AIR@InnoHK, Hong Kong.}

\subsection*{Data availability}
\revision{Data supporting the findings of this study are available from the corresponding author upon reasonable request.}

\bibliography{sn-bibliography}

\end{document}